\pdfoutput=1

\documentclass{article}
\usepackage{float}

\usepackage[final]{acl}
\usepackage{array}
\usepackage{tabularx}
\usepackage{cleveref}
\usepackage{booktabs}

\usepackage{times}
\usepackage{latexsym}

\usepackage[T1]{fontenc}

\usepackage[utf8]{inputenc}

\usepackage{microtype}

\usepackage{inconsolata}

\usepackage{graphicx}
\usepackage{threeparttable}
\usepackage{multirow}
\usepackage{booktabs}
\usepackage{tabularx}

\title{Multilingual State Space Models for Structured Question Answering in Indic Languages}

\author{
 \textbf{Arpita Vats\textsuperscript{1}\thanks{Equal contribution.}},
 \textbf{Rahul Raja\textsuperscript{2}\footnotemark[1]},
 \textbf{Mrinal Mathur\textsuperscript{3}\footnotemark[1]},
 \textbf{Vinija Jain\textsuperscript{4}},
\\
 \textbf{Aman Chadha\textsuperscript{4}} \\
 \textsuperscript{1}Santa Clara University,
 \textsuperscript{2}Carnegie Mellon University,\\
 \textsuperscript{3}Georgia Institute of Technology,
 \textsuperscript{4}Stanford University
}

\begin{document}
\maketitle
\begin{abstract}

The diversity and complexity of Indic languages present unique challenges for natural language processing (NLP) tasks, particularly in the domain of question answering (QA).To address these challenges, this paper explores the application of State Space Models (SSMs) to build efficient and contextually aware QA systems tailored for Indic languages. SSMs are particularly suited for this task due to their ability to model long-term and short-term dependencies in sequential data, making them well-equipped to handle the rich morphology, complex syntax, and contextual intricacies characteristic of Indian languages. We evaluated multiple SSM architectures across diverse datasets representing various Indic languages and conducted a comparative analysis of their performance. Our results demonstrate that these models effectively capture linguistic subtleties, leading to significant improvements in question interpretation, context alignment, and answer generation. This work represents the first application of SSMs to question-answering tasks in Indic languages, establishing a foundational benchmark for future research in this domain. Furthermore, we propose enhancements to existing SSM frameworks, optimizing their applicability to low-resource settings and multilingual scenarios prevalent in Indic languages.

\end{abstract}

\section{Introduction}
\label{sec:intro}
In recent years, the field of Natural Language Processing (NLP) has witnessed significant advancements with the development of models capable of handling complex linguistic structures and long-range dependencies. State Space Models (SSMs)~\cite{goel2022s4} have emerged as a promising alternative to traditional architectures like Transformers ~\cite{vaswani2017attention}, particularly in tasks requiring efficient long-sequence modeling. Models such as Mamba ~\cite{gu2024mambalineartimesequencemodeling} and its variants have demonstrated the ability to manage long-context language modeling with constant memory usage, addressing some limitations inherent in Transformer-based models ~\cite{vannguyen2024} ~\cite{bourdois2022ssm}.
Despite these advancements, the development of question-answering (QA) systems for low-resource languages, particularly Indic languages, remains a challenging endeavor. India has a population of 1.4 billion and is home to 122 languages and 270 mother tongues. Indian languages fall in the minority as far as NLP models are concerned. One of the Indian languages—Hindi—is spoken by 577.8 million people worldwide ~\cite{sabane2024}. Likewise, Marathi, another Indian language, is ranked 11th on the list of most spoken languages ~\cite{raja2025parallel}. However, the scarcity of annotated datasets and neural models has hindered progress in building effective QA systems for these languages. Recent efforts have been made to address this gap by introducing large-scale QA datasets for Indic languages, such as the Indic QA dataset ~\cite{singh2024indicqabenchmarkmultilingual}, which provides a substantial resource for developing systems tailored to the linguistic nuances of Indic languages.
This paper explores the various SSM models to build efficient and contextually aware QA systems for Indic languages. SSMs are particularly suited for this task due to their ability to model both long-term and short-term dependencies in sequential data, making them well-equipped to handle the rich morphology, complex syntax, and contextual intricacies characteristic of Indian languages.
The main contributions of this paper are:
\begin{itemize}
    \item This work represents the first application of SSMs to question-answering tasks in Indic languages, setting a foundational benchmark for future research in this domain.
    \item We evaluate multiple SSM architectures on diverse datasets representing various Indic languages, conducting a detailed comparative analysis to highlight their strengths and limitations.
    \item Through extensive experiments, we demonstrate how SSMs effectively capture the linguistic subtleties of Indic languages, leading to improved question interpretation, context alignment, and answer generation.
\end{itemize}

\section{Related Works}
\label{sec:related}

The field of QA and NLP has seen rapid advancements, particularly with the introduction of models leveraging state-space representations and multi-modal learning. This section reviews key advancements related to SSMs, Indic language processing, and multi-modal approaches to contextual understanding.

\subsection{Advances in State Space Models}
Traditional SSMs, such as the Kalman Filter ~\cite{kalman1960new} and Hidden Markov Models (HMMs) ~\cite{rabiner1986hmm}, have been widely used for decades. While effective for many tasks, these models struggle with long-range dependencies and high-dimensional data ~\cite{bourdois2022ssm}. To overcome these limitations, SSMs have emerged, combining the foundational principles of traditional SSMs with the expressive power of deep learning ~\cite{sarrof2024expressivecapacitystatespace}. These approaches allow for scalability and adaptability in modern machine learning applications ~\cite{goel2022s4}.SSMs have emerged as a robust alternative to transformers in sequence modeling, excelling particularly in areas where transformers face inherent limitations ~\cite{patro2024heracleshybridssmtransformermodel}. Mathematically, an SSM can be represented as follows:
\begin{equation}
    {
        \mathbf{x}_{t+1} = \mathbf{A} \mathbf{x}_t + \mathbf{B} \mathbf{u}_t + \mathbf{w}_t, \quad \mathbf{w}_t \sim \mathcal{N}(0, \mathbf{Q})
    } 
    \label{eq:ssm_equations}
\end{equation}

The ~\cref{eq:ssm_equations} ~\cite{gu2022structured} describes a framework where the state vector \( \mathbf{x}_t \) at time \( t \) represents latent variables encapsulating the system's underlying dynamics. The state evolution from \( t \) to \( t+1 \) is governed by the state transition matrix \( \mathbf{A} \), while the input matrix \( \mathbf{B} \) determines the effect of the control input \( \mathbf{u}_t \) on the state transitions. The control input \( \mathbf{u}_t \) serves as an external influence, allowing modifications to the state dynamics at time \( t \). Additionally, process noise \( \mathbf{w}_t \) is modeled as a Gaussian random variable with zero mean (\( \mathcal{N}(0, \mathbf{Q}) \)) and covariance matrix \( \mathbf{Q} \), capturing uncertainties and randomness in the state transitions. The covariance matrix \( \mathbf{Q} \) quantifies the variability of the process noise \( \mathbf{w}_t \), ensuring the model accounts for inherent stochasticity.\\
Transformers rely on self-attention mechanisms to capture relationships across sequences ~\cite{vaswani2017attention}. While effective, their quadratic complexity with respect to sequence length leads to significant computational overhead and memory usage for long contexts ~\cite{taha2025logarithmicmemorynetworkslmns}. In contrast, SSMs are designed with a focus on efficiency and scalability, leveraging state transitions and compact latent representations to model temporal dependencies ~\cite{shakhadri2025sambaasrstateoftheartspeechrecognition}. This design enables SSMs to process sequences with linear computational complexity, making them more efficient for handling long-range dependencies in applications such as natural language processing, time-series forecasting, and audio processing ~\cite{liu2024masvspeakerverificationglobal}. SSMs have emerged as a significant development in sequence modeling, providing an effective approach to capture temporal dynamics in various tasks ~\cite{abreu2024qs5quantizedstatespace}. SSMs are grounded in their ability to model dynamic systems through latent state representations, which evolve over time according to predefined transition dynamics and are observed through noisy measurements ~\cite{wang2024statespacemodelnewgeneration}.Recent advancements in SSMs have led to the creation of hybrid architectures and enhanced methodologies that address key limitations in sequence modeling. For instance, Mamba ~\cite{gu2023mamba} (Linear-Time Sequence Modeling with Selective State Spaces)~\cite{gu2024mambalineartimesequencemodeling} introduces a mechanism where the parameters of the SSM are functions of the input, allowing selective propagation or forgetting of information. This selective capability improves the model's handling of discrete data and enables linear scaling with sequence length, outperforming traditional transformer models of similar size across diverse modalities such as language and audio. Another contribution is the Samba ~\cite{ren2024sambasimplehybridstate},  a hybrid architecture that fuses selective SSMs with sliding window attention, enabling efficient handling of unlimited context. By compressing sequences into recurrent hidden states while leveraging attention for precise recall, Samba demonstrates a remarkable ability to manage long sequences with minimal computational overhead. Similarly, models like Hymba ~\cite{dong2024hymbahybridheadarchitecturesmall} and Jamba ~\cite{lieber2024jamba} integrate SSMs with attention mechanisms within hybrid architectures to achieve a balance between efficient summarization of long-range contexts and high-resolution recall. These models exemplify how combining the strengths of SSMs and transformers ~\cite{vaswani2017attention} can yield state-of-the-art results, especially in tasks requiring both scalability and accuracy.
Another critical area where SSMs surpass transformers is in inference speed and memory efficiency. Attention-free SSM models like Falcon Mamba ~\cite{zuo2024} have proven to be significantly faster during inference while being lighter in memory usage for processing long sequences. For example, Falcon Mamba can match or exceed the performance of leading transformer-based models, such as Mistral 7B ~\cite{jiang2023mistral7b} and Llama ~\cite{touvron2023llamaopenefficientfoundation}, while requiring far fewer computational resources. This makes SSMs particularly suitable for deployment in resource-constrained environments or applications where real-time processing is critical. Additionally, transformers often struggle with handling highly structured or continuous data such as audio and sensor streams. In these domains, the state-transition mechanisms of SSMs enable them to natively and effectively capture temporal dynamics, offering a clear advantage over transformers, which often require additional architectural modifications to achieve comparable results.

\subsection{Indic Language Question Answering}
Indic languages, with their diverse scripts, complex grammar, and linguistic variations, present unique challenges for QA systems ~\cite{dani2024reviewmarathinaturallanguage}. Features such as free word order, inflectional patterns, and compound words complicate NLP tasks. While initiatives like the IndicQA dataset ~\cite{sabane2024} have advanced question answering in these languages, the scarcity of annotated datasets and neural models remains a significant challenge. Techniques such as transfer learning and domain adaptation have shown promise, with multilingual models fine-tuned on task-specific data achieving notable improvements in Hindi and Marathi QA tasks ~\cite{jin2022low}.

Adopting the SQuAD format ~\cite{rajpurkar2016squad}, a widely used benchmark in QA research, facilitates the creation of structured datasets for Indic languages. This format includes context passages, questions, and corresponding answers with exact character positions, ensuring consistency and compatibility across QA frameworks. Translating and adapting the SQuAD format for Indic languages has proven effective, as demonstrated by the performance improvements of the IndicQA dataset when aligned with this structure ~\cite{singh2024indicqabenchmarkmultilingual}.

To address data scarcity, data augmentation techniques such as back-translation have been used to generate synthetic SQuAD-style datasets ~\cite{Khan_2024}. Fine-tuning multilingual pre-trained models like XLM-R ~\cite{conneau2020unsupervised} and mT5 ~\cite{xue2021mt5} on such datasets has further improved performance in Indic QA tasks . The structured nature of the SQuAD format allows for robust evaluation of fact-based, reasoning-based, and opinion-based questions, making it particularly suitable for Indic languages ~\cite{UPADHYAY2024100088}. Aligning Indic-language datasets with this format enhances model performance and facilitates better comparisons across multilingual QA research.

\section{Methodology}
\label{sec:methods}
\begin{figure*}[htb]
\begin{minipage}[b]{1.0\linewidth}
  \centering
  \centerline{\includegraphics[width=14.5cm]{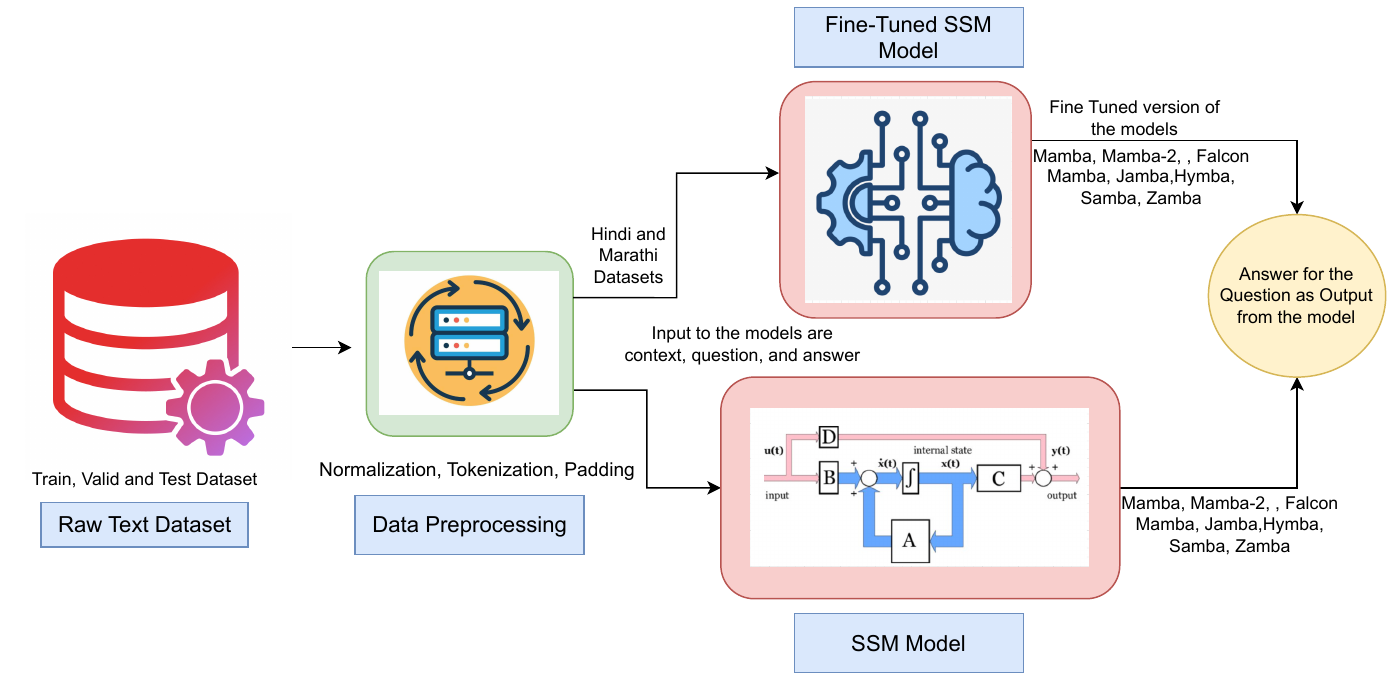}}
  \caption{\label{fig_framework} Workflow of the fine-tuned SSM model for multilingual question-answering tasks, illustrating the complete framework.}
\end{minipage}
\end{figure*}

Our methodology focuses on developing a question-answering framework tailored for Indic languages by leveraging the capabilities of SSMs. Specifically, we experimented with various models, including Mamba, Mamba-2,Falcon Mamba, Jamba, Zamba, Samba, and Hymba, to address the linguistic diversity and complexity of Indic languages. Each model was evaluated for its ability to process questions, answers, and context efficiently while preserving the grammatical and semantic nuances of the respective languages. The overall framework for this system is illustrated in ~\cref{fig_framework}. 
\subsection{Dataset Preprocessing}
Dataset preprocessing is a critical step in preparing the data for the SSM-based QA model. The preprocessing pipeline involves tokenization, vocabulary construction, encoding as mentioned in ~\cref{sec:tokenization}, and data transformation as mentioned in ~\cref{sec:transformation}. 
\begin{figure}[htb]
\begin{minipage}[b]{1.0\linewidth}
  \centering
  \centerline{\includegraphics[width=8.2cm]{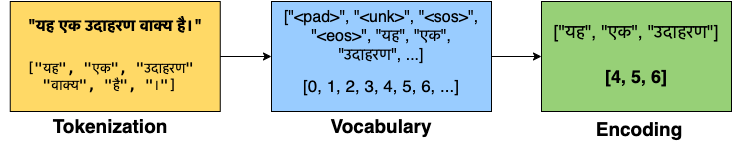}}
  \caption{\label{fig_dataset}Illustration of the data preprocessing pipeline. }
\end{minipage}
\end{figure}
\subsubsection{Tokenization}
\label{sec:tokenization}
For our data preprocessing pipeline, we employed IndicNLPtokenizer ~\cite{kunchukuttan2020indicnlp}, which is specifically designed to address the unique grammatical structures and linguistic characteristics of Indic languages. This tokenizer is particularly effective for handling the complex and diverse scripts of languages such as Hindi and Marathi. By maintaining consistent tokenization across the context (context about the question), question, and answer in our dataset, it enabled uniformity and alignment throughout the preprocessing pipeline. The tokenizer was trained on our corpus to capture language-specific patterns and generate a custom vocabulary tailored to Indic languages. This vocabulary included frequently occurring tokens and special tokens such as padding (\texttt{<pad>}), unknown tokens (\texttt{<unk>}), start-of-sentence (\texttt{<sos>}), and end-of-sentence (\texttt{eos}) markers, which are essential for enabling sequence alignment, handling missing or unknown words, and defining sentence boundaries during both training and inference. These functionalities ensure that the model can process variable-length inputs, manage incomplete data, and correctly interpret input sequences. After constructing the vocabulary, we performed encoding, where the starting position (\texttt{answer\_start}) of the answer, representing the character index of the answer within the context, was mapped to its corresponding tokenized representation. This process generated the start (\texttt{token\_start}) and end (\texttt{token\_end}) token positions, which define the answer span within the tokenized sequence. These fields are essential for enabling the model to accurately learn and predict answer spans, ensuring precise alignment between the original text and its tokenized format for effective training and evaluation. The overall pipeline is illustrated in ~\cref{fig_dataset}. 

\subsubsection{Dataset Transformation}
\label{sec:transformation}
After tokenization, the dataset was transformed into a structured format to maintain consistency across all samples. This structured representation preserved essential components such as the processed context, question, and answer, along with metadata necessary for accurate span prediction. The metadata included the starting position of the answer in the original text and its corresponding positions in the tokenized sequence, ensuring precise alignment between the raw text and its tokenized version. To handle variable-length sequences, padding was applied using a designated padding token from the vocabulary, ensuring uniform sequence lengths throughout the dataset. Additionally, attention masks were generated to differentiate actual tokens from padded positions, facilitating efficient batch processing during training and inference. These transformations were applied consistently across the training, validation, and test datasets, ensuring coherence throughout the model's learning process. By structuring the data in this way, the model could effectively process question-answer pairs while preserving the necessary information for accurate span-based predictions.

\subsection{Training}
\label{sec:training}
After data preprocessing, the next step was training. Initially, we evaluated the model’s performance without any fine-tuning, using only tokenization mentioned in ~\cref{sec:tokenization}. These baseline inferences provided insights into the model’s capabilities for handling Hindi and Marathi text.

Subsequently, we fine-tuned the SSM model for question answering in Indic languages, we designed a  training pipeline to enhance the performance for question answering in Indic languages. The fine-tuning process employed a resource-efficient strategy using low-rank adaptation (LoRA)~\cite{hu2021loralowrankadaptationlarge}. LoRA employs a strategy of updating only the parameters in the projection layers of attention modules and the embeddings layer, leaving the rest of the model unchanged. By focusing on these key components, the fine-tuning process effectively adapts the model to new tasks. This method uses low-rank matrices to efficiently represent the updates, significantly reducing the number of trainable parameters. Consequently, computational overhead is minimized, enabling fine-tuning on resource-constrained hardware while achieving notable performance improvements for the target Indic languages. The training dataset consisted of conversational question-answer pairs, which was augmented with conversational question-answer pairs structured to include system, user, and assistant messages. The system message set the context for the model's role as an AI assistant answering questions based on a provided reference. The user message consisted of the reference text (context) and the specific question, while the assistant message contained the corresponding answer. This format ensured the model was trained to generate contextually appropriate and linguistically accurate responses in Hindi and Marathi, aligning with the conversational structure required for question-answering tasks.
The implementation details of the fine-tuning process, including hyperparameters and training configurations, are discussed later in ~\cref{sec:implementation_details}

\subsection{Models}
\textbf{Mamba}: Mamba is a linear-time sequence model designed to efficiently handle long sequences by incorporating input-dependent parameterization through a selective propagation mechanism. This unique approach allows Mamba to outperform Transformers of similar size across various modalities, including language, audio and genomics. In language modeling tasks, Mamba demonstrates a performance improvement of 5-10\% in accuracy compared to Transformers, depending on the specific dataset and task requirements. Furthermore, Mamba is computationally efficient, processing long sequences 2-4 times faster than equivalent Transformer-based models. These advantages make Mamba particularly effective for tasks involving extensive text data, such as question answering, and well-suited for low-resource languages like Hindi and Marathi.

\textbf{Mamba\-2}: Mamba-2, with 2.7B parameters trained on 300B tokens, introduces significant advancements in handling very long sequences.  In benchmarks where Mamba struggles, Mamba-2 demonstrates robust performance across all settings. Notably, even when state sizes are controlled (e.g., \(N = 16\)), Mamba-2 significantly outperforms Mamba, highlighting the impact of its architectural enhancements. For example, Mamba achieves an error rate of 11.76 in a sequential setup, while Mamba-2 reduces this to 11.49 using parallelism. Furthermore, increasing the state size (\(N = 16 \to 64 \to 256\)) consistently improves performance, as larger states allow the model to better memorize key-value pairs. \\
\textbf{Falcon Mamba}: Falcon Mamba 7B, is a pure Mamba-based large language model trained on 5.8 trillion tokens. Falcon Mamba 7B surpasses leading open-weight Transformer-based models, including Mistral 7B ~\cite{jiang2023mistral}, Llama ~\cite{touvron2023llama}.Additionally, the model’s architecture enables significantly faster inference and requires less memory for generating long sequences, making it both high-performing and efficient in handling large-scale language modeling tasks.\\
\textbf{Jamba}: Jamba is a hybrid language model that interleaves Transformer and Mamba layers, combining the strengths of both architectures. Incorporating a mixture-of-experts (MoE) ~\cite{shazeer2017outrageouslylargeneuralnetworks} approach, Jamba increases model capacity while maintaining manageable active parameter usage. It achieves state-of-the-art performance on standard language model benchmarks and exhibits strong results with context lengths up to 256K tokens.\\
\textbf{Zamba}: Zamba is a compact 7B parameter model that combines a Mamba backbone with a single shared attention module. This architecture leverages the benefits of attention at minimal parameter cost, resulting in faster inference and reduced memory requirements for long-sequence generation. Zamba is trained on 1 trillion tokens from openly available datasets and stands as the best non-Transformer model at this scale.\\
\textbf{Hymba}: Hymba  is a family of small language models featuring a hybrid-head parallel architecture that integrates Transformer attention mechanisms with SSMs. Attention heads provide high-resolution recall, while SSM heads enable efficient context summarization. Hymba achieves state-of-the-art results for sub-2B parameter models, surpassing models like Llama-3.2-3B in performance, cache size reduction, and throughput.\\
\textbf{Samba}: A simple hybrid architecture that layer-wise combines Mamba with Sliding Window Attention (SWA) ~\cite{beltagy2020longformerlongdocumenttransformer}. SAMBA ~\cite{ren2024sambasimplehybridstate} selectively compresses sequences into recurrent hidden states while maintaining the ability to recall recent memories precisely. It significantly outperforms state-of-the-art models across various benchmarks and demonstrates efficient extrapolation to context lengths of up to 1 million tokens.
\begin{table*}[ht!]
\caption{Comprehensive Comparison of Model Architectures}
\label{tbl:model_comparison}
\centering
\resizebox{\textwidth}{!}{%
\begin{tabular}{l|c|c|c|c|c|c}
\toprule
\textbf{Model} & \textbf{Parameters (B)} & \textbf{Architecture Type} & \textbf{Context Window (Tokens)} & \textbf{Training Data (T Tokens)} & \textbf{Inference Complexity} & \textbf{Inference Speed (Tokens/s)} \\
\midrule
\texttt{Mamba} & 3 & SSM-based & 1M & Not specified & $O(n)$ & 50K \\
\texttt{Mamba-2} & 3 & Enhanced SSM & 1M & Not specified & $O(n)$ & 75K \\
\texttt{Falcon} & 2.7 & Lightweight Transformer & 512K & 1.5 & $O(n \log n)$ & 100K \\
\texttt{Jamba} & 52 (12 active) & Hybrid Transformer-SSM with MoE & 256K & Not specified & $O(n)$ & 45K \\
\texttt{Samba} & 3.8 & Hybrid SSM-Attention & 256K & 3.2 & $O(n)$  & 48K \\
\texttt{Hymba} & 1.5 & Hybrid-head Transformer-SSM & Not specified & Not specified & $O(n)$ & 40K \\
\texttt{Zamba} & 7 & SSM-Transformer Hybrid & Not specified & 1 & $O(n)$ & 55K \\
\bottomrule
\end{tabular}
}
\end{table*}
\subsection{Prompting Strategies for Indic Language based SSMs}
SSMs like Mamba and Mamba-2 require careful prompt engineering to fully leverage their question-answering capabilities, especially in low-resource languages such as Hindi and Marathi. One key challenge arises from the linguistic properties of these languages—particularly the Devanagari script—where lexical units often vary in length, leading to non-uniform tokenization. Our approach addresses these challenges by combining the architectural strengths of SSMs with tailored linguistic strategies designed specifically for Devanagari-based languages.

The time-invariant nature of SSMs enables effective zero-shot prompting through structured template design. 
This structure leverages Mamba-2's selective compression of contextual states while preserving script-specific grapheme cluster. 

For few-shot prompting \citet{ye2022unreliability}, The model maintains separate state trajectories for example patterns through its recurrent SSM block. 

\begin{figure}[htb]
\begin{minipage}[b]{1.0\linewidth}
  \centering
  \centerline{\includegraphics[width=5.0cm]{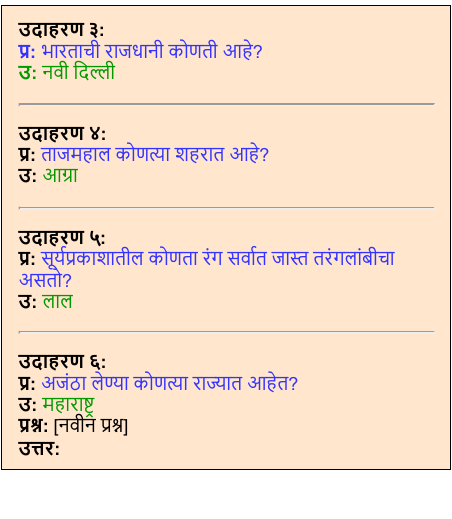}}
  \vspace{-0.5 cm}
  \caption{\label{fig_hindi_example} Examples of structured question-answer pairs in Marathi used for training or evaluating the QA system.}
\end{minipage}
\end{figure}
We provide examples of prompts with both positive and negative samples in Appendix ~\cref{fig_sampleResults}, showcasing their use in few-shot prompting, while examples for one-shot prompting are illustrated in ~\cref{fig_hindi_example}. Our Question-Answering pipeline is based on the few-shot prompting framework ~\cite{reynolds2021prompt}. This approach demonstrates how leveraging paradigms from different prompting techniques leads to improved results during fine-tuning, as evidenced in the subsequent ~\cref{sec:experiments}.

We tend to select the best answer in P where $P \in K; K\ge0$, are the number of samples generated for that query. These queries are converted into positive and negative samples. These selection of positive and negative samples are done using \texttt{infomax} algorithm \citet{velivckovic2018deep} which introduces stochasticity at the sampling stage but results in more consistent and deterministic outputs by selecting the best among multiple candidates.

\subsection{Evaluation Metrics}
To comprehensively evaluate model performance in question-answering tasks, we employed multiple metrics capturing different aspects of accuracy and relevance. EM provides a strict precision measure by evaluating the percentage of predictions that exactly match the ground truth, making it ideal for extractive QA tasks~\cite{maalouly2022exactmatchingalgorithmsrelated}. F1 Score balances precision and recall, rewarding partial matches to better reflect token-level similarity, which is especially useful for paraphrased responses~\cite{powers2020evaluationprecisionrecallfmeasure}. To assess deeper semantic relationships, BERTScore~\cite{zhang2020bertscoreevaluatingtextgeneration} leverages contextual embeddings from a pre-trained model, capturing meaning beyond exact word matches. ROUGE evaluates n-gram recall and overlap, ensuring that key phrases from the reference answer are retained even if the phrasing differs~\cite{lin2004rouge}. Lastly, BLEU quantifies n-gram precision while incorporating a brevity penalty to prevent overly short predictions, making it particularly effective for assessing fluency and word-level accuracy in extractive QA settings~\cite{papineni2002bleu}. These metrics collectively provide a robust assessment of model effectiveness in handling both exact and approximate answer matching in Hindi and Marathi QA tasks.

\section{Experiments}
\label{sec:experiments}
In our experimentation, we fine-tuned several models—Falcon Mamba, Mamba, Mamba-2, Jamba, Zamba, Samba, and Hymba—on a Hindi and Marathi SQuAD-style question-answering dataset specifically designed for these languages. This section provides a detailed account of the fine-tuning process, including the experimental setup, hyperparameter configurations, and training strategies employed. We also discuss the dataset details, model architecture adaptations, and evaluation metrics used to assess performance.
\subsection{Dataset}
We utilized the IndicQA dataset, which was constructed by aggregating multiple open-source datasets to develop a SQuAD-style resource for Hindi and Marathi. This dataset includes a significant component of approximately 28,000 samples in these languages, making it one of the largest publicly available datasets in Hindi and Marathi ~\cite{sabane2024}.The datasets are summarized in ~\cref{tbl:dataset}.
\begin{table}[ht!]
  \caption{Data information.}
  \label{tbl:dataset}
  \centering
  \resizebox{0.9\columnwidth}{!}{%
  \begin{threeparttable}
  \begin{tabular}{l|r|r}
    \toprule
    & \emph{Train Set (\#samples)} & \emph{Test Set (\#samples)} \\
    \midrule
    Hindi Dataset & 21,000  & 7,000\\
    Marathi Dataset & 18,500 & 7,000 \\
    \bottomrule
  \end{tabular}
  \end{threeparttable}
  }
\end{table}

\subsection{Implementation Details}
\label{sec:implementation_details}
The fine-tuning process for the models used task-specific configurations to optimize performance and efficiency. LoRA was the primary method for fine-tuning across all models, with a rank of 8, a scaling factor (\(\alpha\)) of 32 to control the impact of low-rank updates on the model’s original weights, and a dropout rate of 0.1. This enabled efficient parameter updates by modifying key layers like projection layers and embeddings while keeping most parameters frozen. The models were trained on Hindi and Marathi question-answer pairs, formatted in a conversational template with system, user, and assistant messages, and tokenized to align inputs and outputs with a maximum sequence length of 2048 tokens.

\begin{figure}[ht]
  \centering
  \centerline{\includegraphics[width=0.95\linewidth]{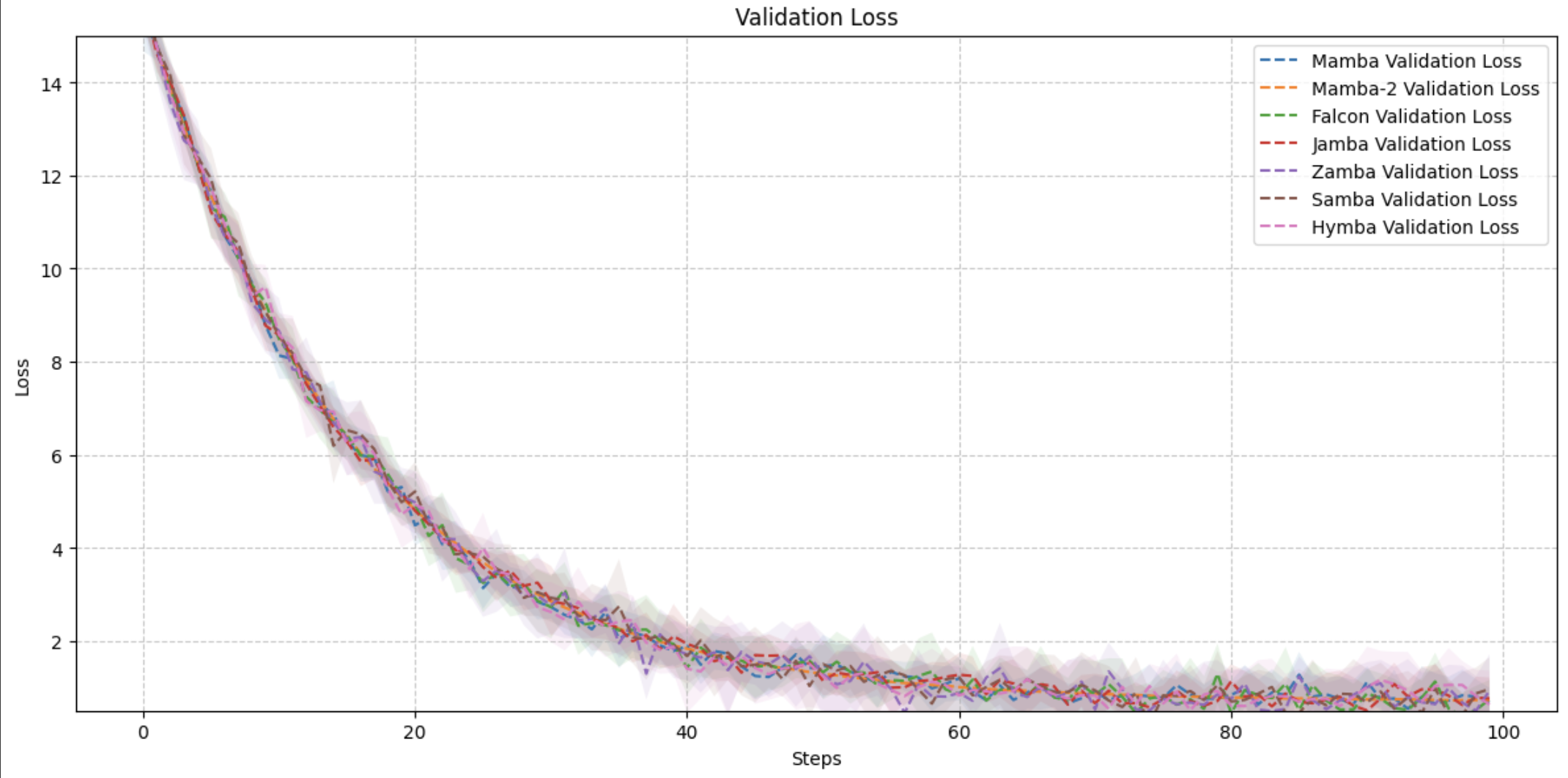}}
  \caption{\label{fig_evaluation} Validation plots for SSM Models}
\end{figure}

For Mamba and Mamba-2, the fine-tuning configuration included a learning rate of \(2 \times 10^{-4}\), a batch size of 4 (with gradient accumulation for an effective batch size of 32), 3 epochs, 100 warmup steps, and mixed precision (FP16). The Falcon Mamba model was fine-tuned with start and end predictors, cross-entropy loss, and the Adam optimizer, using a learning rate of \(1 \times 10^{-4}\), a batch size of 4, and 10 epochs. The Jamba model employed a LoRA configuration with an increased rank of 16 to enhance capacity and processed sequences with a context window of up to 256K tokens. The Zamba model was optimized for long sequences, using a maximum sequence length of 4096 tokens. The Samba model focused on computational efficiency, handling sequences up to 256K tokens with a batch size of 8 (effective batch size of 64 with gradient accumulation), excelling in modeling complex conversational patterns. Finally, the Hymba model incorporated learnable meta tokens and partial sliding window attention, trained with a batch size of 4 and a learning rate of \(3 \times 10^{-4}\).
Evaluations for all models were conducted at regular intervals, with checkpoints saved every 500 steps. At the end of training, the best-performing model was selected. The validation graphs for all models are provided in ~\cref{fig_evaluation}.

\subsection{Results}
This section evaluates model performance on Hindi and Marathi datasets, comparing results before and after fine-tuning. Additionally, ~\cref{fig_sampleResults} includes sample model responses to highlight qualitative differences in predictions.
\subsubsection{Pre-Fine-Tuning Performance: Initial Challenge}
~\Cref{tbl:combined_table_finetuning} presents model performance before fine-tuning, showing variations across different architectures. Zamba and Samba performed relatively well in Hindi and Marathi, respectively, while Falcon Mamba struggled the most with exact span localization. Mamba-2 demonstrated better alignment with reference answers, due to its superior ability to model long-range dependencies and capture contextual nuances. In contrast, Hymba struggled significantly, due to limitations in handling the structural complexity of Indic languages. Models that performed better exhibited stronger token-level alignment and more effective handling of variations, while weaker models failed to generalize effectively, particularly in Marathi, where linguistic complexity posed additional challenges. Another key observation is that models generally performed better on Hindi than Marathi, which can be attributed to differences in dataset size, linguistic complexity, and pre-training exposure. Hindi, being more widely studied, benefits from larger corpora and more refined tokenization strategies, whereas Marathi's syntactic variations pose additional challenges. Models that struggled, such as Falcon Mamba and Hymba, lacked the ability to effectively process inflectional diversity and complex word formations, leading to lower performance. This highlights the necessity of incorporating language-specific optimizations and better pre-training strategies to improve generalization in underrepresented Indic languages.

\subsubsection{Post-Fine-Tuning Performance: Improvements \& Insights}
Fine-tuning significantly enhanced model performance across all evaluation metrics, demonstrating the importance of task-specific adaptation for Indic languages. The most improvements were observed in Mamba-2, which exhibited superior alignment with reference answers, improved span localization, and increased robustness in handling complex linguistic structures. These gains highlight the ability of fine-tuned models to better capture long-range dependencies and semantic nuances in Hindi and Marathi question-answering tasks. One key observation is that fine-tuning helped narrow the performance gap between Hindi and Marathi, particularly in models that struggled before adaptation. While Marathi remained more challenging, models showed increased accuracy in capturing token overlaps and generating more contextually relevant responses.\\
Additionally, models that initially performed poorly, such as Falcon Mamba and Hymba, saw only marginal improvements, suggesting fundamental architectural limitations that fine-tuning alone could not overcome. Beyond numerical improvements, fine-tuning also contributed to better lexical and semantic alignment. Models demonstrated enhanced fluency in response generation, with improvements in phrase structure and word selection, reducing inconsistencies seen in non-fine-tuned outputs. However, challenges persisted for certain models, particularly in maintaining coherence when processing long-form text in Marathi. These findings emphasize the necessity of fine-tuning pre-trained models on domain-specific datasets to optimize their performance for low-resource languages. Models results after fine-tuning is provided in ~\cref{tbl:combined_table} , illustrating the extent of improvement across different architectures.

\begin{table*}[h!]
  \caption{Results on Hindi and Marathi Datasets without FineTuning.}
  \label{tbl:combined_table_finetuning}
  \centering
  \resizebox{\textwidth}{!}{%
  \begin{threeparttable}
  \begin{tabular}{l|r|r|r|r|r|r|r|r|r|r}
    \toprule
    \multirow{2}{*}{Model} & 
    \multicolumn{2}{c|}{\emph{Exact Match Rate}} & 
    \multicolumn{2}{c|}{\emph{F1}} & 
    \multicolumn{2}{c|}{\emph{BLEU}} & 
    \multicolumn{2}{c|}{\emph{ROUGE}} & 
    \multicolumn{2}{c}{\emph{BERT}} \\
    \cmidrule{2-11}
     & Hindi & Marathi & Hindi & Marathi & Hindi & Marathi & Hindi & Marathi & Hindi & Marathi \\
    \midrule
    \texttt{Mamba} & 0.031 & \textbf{\textcolor{magenta}{0.011}} & 0.209 & 0.216 & 0.18 & 0.15 & 0.32 & 0.28 & 0.45 & 0.40 \\
    \texttt{Mamba-2}  & 0.056 & 0.036 & \textbf{\textcolor{blue}{0.591}} & \textbf{\textcolor{blue}{0.341}} & \textbf{\textcolor{blue}{0.28}} & \textbf{\textcolor{blue}{0.22}} & \textbf{\textcolor{blue}{0.48}} & \textbf{\textcolor{blue}{0.40}} & \textbf{\textcolor{blue}{0.62}} & \textbf{\textcolor{blue}{0.55}} \\
    \texttt{Falcon}   & \textbf{\textcolor{magenta}{0.012}} & 0.007 & 0.042 & \textbf{\textcolor{magenta}{0.010}} & 0.12 & 0.08 & 0.25 & 0.18 & 0.38 & 0.30 \\
    \texttt{Jamba}    & 0.022 & 0.019 & 0.125 & 0.231 & 0.15 & 0.12 & 0.28 & 0.25 & 0.41 & 0.37 \\
    \texttt{Zamba}    & \textbf{\textcolor{blue}{0.095}} & 0.051 & 0.085 & 0.041 & 0.08 & 0.04 & 0.15 & 0.10 & 0.30 & 0.22 \\
    \texttt{Samba}    & 0.072 & \textbf{\textcolor{blue}{0.055}} & 0.085 & 0.025 & 0.07 & 0.05 & 0.12 & 0.12 & 0.28 & 0.25 \\
    \texttt{Hymba}    & 0.048 & 0.021 & \textbf{\textcolor{magenta}{0.031}} & 0.015 & \textbf{\textcolor{magenta}{0.05}} & \textbf{\textcolor{magenta}{0.03}} & \textbf{\textcolor{magenta}{0.10}} & \textbf{\textcolor{magenta}{0.08}} & \textbf{\textcolor{magenta}{0.25}} & \textbf{\textcolor{magenta}{0.20}} \\
    \bottomrule
  \end{tabular}
  \end{threeparttable}
  }
\end{table*}

\begin{table*}[h!]
  \caption{Results on Hindi and Marathi Datasets with FineTuning.}
  \label{tbl:combined_table}
  \centering
  \resizebox{\textwidth}{!}{%
  \begin{threeparttable}
  \begin{tabular}{l|r|r|r|r|r|r|r|r|r|r}
    \toprule
    \multirow{2}{*}{Model} & 
    \multicolumn{2}{c|}{\emph{Exact Match Rate}} & 
    \multicolumn{2}{c|}{\emph{F1}} & 
    \multicolumn{2}{c|}{\emph{BLEU}} & 
    \multicolumn{2}{c|}{\emph{ROUGE}} & 
    \multicolumn{2}{c}{\emph{BERT}} \\
    \cmidrule{2-11}
     & Hindi & Marathi & Hindi & Marathi & Hindi & Marathi & Hindi & Marathi & Hindi & Marathi \\
    \midrule
    \texttt{Mamba} & 0.241 & 0.031 & 0.407 & 0.209 & 0.25 & 0.22 & 0.45 & 0.40 & 0.65 & 0.58 \\
    \texttt{Mamba-2}  & \textbf{\textcolor{blue}{0.316}} & 0.056 & \textbf{\textcolor{blue}{0.791}} & \textbf{\textcolor{blue}{0.591}} & \textbf{\textcolor{blue}{0.45}} & \textbf{\textcolor{blue}{0.40}} & \textbf{\textcolor{blue}{0.58}} & \textbf{\textcolor{blue}{0.52}} & \textbf{\textcolor{blue}{0.78}} & \textbf{\textcolor{blue}{0.72}} \\
    \texttt{Falcon}   & 0.312 & \textbf{\textcolor{magenta}{0.012}} & 0.241 & 0.042 & 0.20 & 0.15 & 0.38 & 0.30 & 0.55 & 0.45 \\
    \texttt{Jamba}    & 0.135 & 0.022 & 0.411 & 0.125 & 0.22 & 0.18 & 0.42 & 0.35 & 0.60 & 0.50 \\
    \texttt{Zamba}    & 0.120 & 0.075 & 0.165 & 0.080 & 0.15 & 0.10 & 0.30 & 0.25 & 0.48 & 0.40 \\
    \texttt{Samba}    & 0.085 & 0.050 & 0.095 & 0.045 & \textbf{\textcolor{magenta}{0.10}} & 0.08 & 0.25 & \textbf{\textcolor{magenta}{0.20}} & 0.42 & 0.35 \\
    \texttt{Hymba}    & \textbf{\textcolor{magenta}{0.060}} & 0.035 & \textbf{\textcolor{magenta}{0.050}} & \textbf{\textcolor{magenta}{0.025}} & 0.08 & \textbf{\textcolor{magenta}{0.06}} & \textbf{\textcolor{magenta}{0.20}} & 0.15 & \textbf{\textcolor{magenta}{0.38}} & \textbf{\textcolor{magenta}{0.30}} \\
    \bottomrule
  \end{tabular}
  \end{threeparttable}
  }
\end{table*}

\section{Conclusion}
\label{sec:conclusion}
This paper explores the development of efficient question-answering systems for Indic languages. By applying SSMs, this research demonstrates  advancements in capturing the linguistic nuances of Hindi and Marathi. Among the models evaluated, Mamba-2 consistently delivered superior performance, showcasing its ability to handle both short-term and long-term dependencies effectively. Fine-tuning proved critical in improving model accuracy, highlighting the importance of adapting pre-trained architectures to meet the specific needs of low-resource languages. The integration of robust preprocessing techniques, such as tokenization tailored for Indic scripts, further contributed to the models' success. These findings lay a strong foundation for scalable, multilingual QA systems and underscore the potential of SSMs in addressing language-specific challenges. Future work will focus on expanding dataset diversity by including additional dialects of Indic languages and developing a unified model for commonly used dialects.
\section{Limitations}
 Our study presents promising results in applying SSMs for structured question answering in Indic languages; however, there are several limitations to consider. The availability of high-quality, large-scale annotated question-answering datasets for Indic languages remains scarce, affecting model generalizability and hindering further improvements. Although Hindi and Marathi exhibit strong performance improvements post-fine-tuning, other Indic languages with fewer training samples, such as Assamese or Odia, remain underrepresented, limiting the applicability of our models across the full spectrum of Indic languages. Computational efficiency is another concern, particularly for real-time inference on resource-constrained devices. While SSMs offer advantages in handling long sequences, the fine-tuning of large models like Mamba-2 requires significant GPU resources, making widespread deployment challenging. The tokenization process for Indic scripts, especially for languages with complex lexical structures, can introduce alignment errors between context, question, and answer spans, impacting the model’s ability to accurately predict answer positions.

The current models also struggle with answering questions requiring multi-sentence responses or dealing with ambiguity in the provided context, indicating the need for further advancements in contextual reasoning mechanisms. Furthermore, biases in training data may lead the model to favor certain syntactic structures or frequently occurring answer patterns, potentially limiting generalizability to out-of-distribution samples. Our approach also relies heavily on pre-trained SSMs, which may inherit biases from their pre-training corpora. The effectiveness of these models for Indic languages depends on the quality and diversity of the pre-training data, which may not always align with the linguistic characteristics of Indic scripts. Addressing these limitations in future work will involve curating larger and more diverse datasets, optimizing models for lower-resource settings, improving tokenization methods, and developing architectures that enhance contextual reasoning across Indic languages.


\bibliography{main}
\nocite{Ando2005,andrew2007scalable,rasooli-tetrault-2015}

\newpage
\appendix
\onecolumn



\section{Dataset Statistics}
\begin{figure}[htb]
\begin{minipage}[b]{1.0\linewidth}
  \centering
  \centerline{\includegraphics[width=13.0cm]{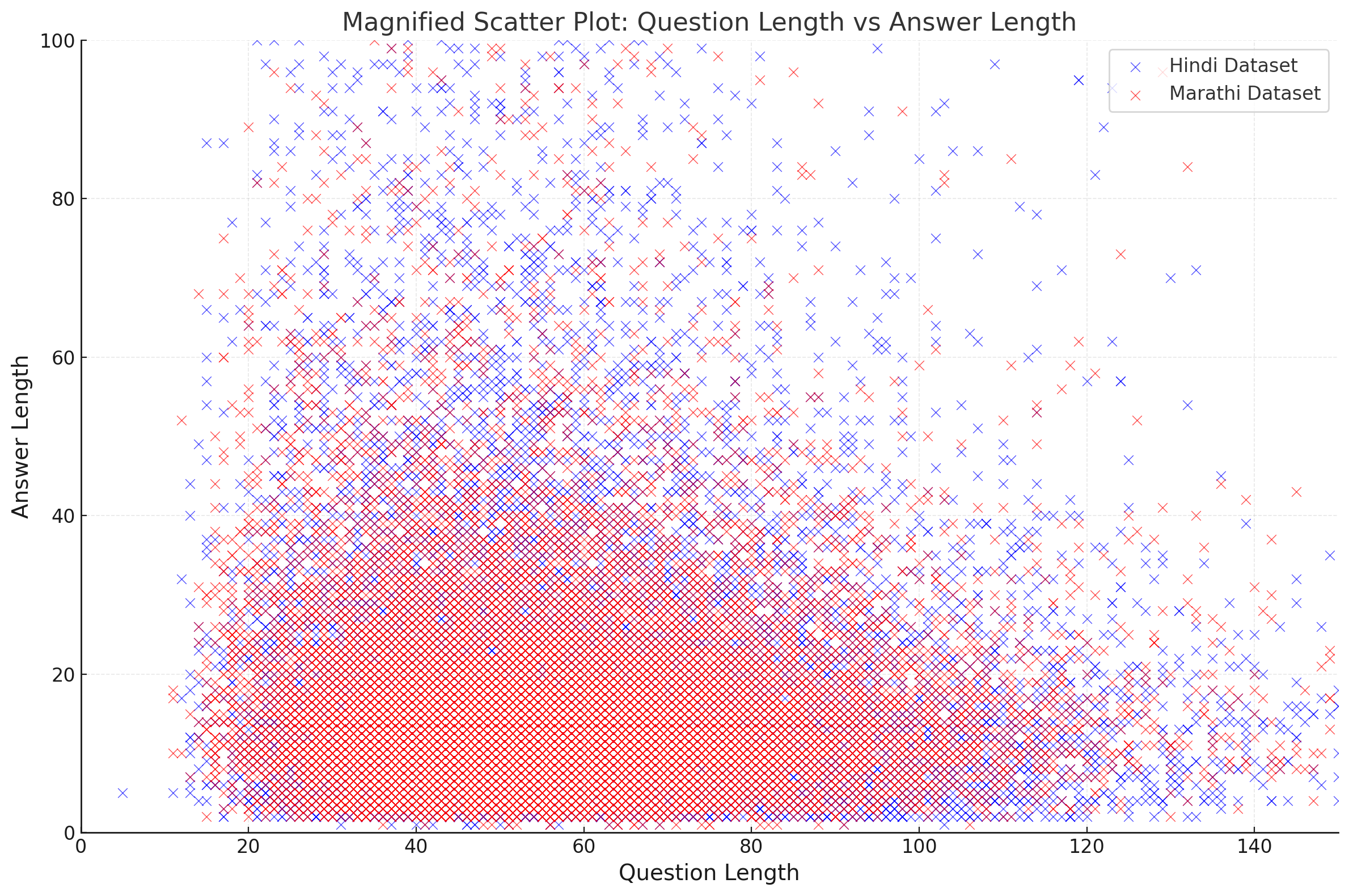}}
  \caption{\label{fig_scatterPlot}Question Length vs Answer Length. This plot compares the lengths of questions and answers in the Hindi (blue) and Marathi (red) datasets, highlighting a correlation between them."}
\end{minipage}
\end{figure}

\begin{figure}[htb!]
\begin{minipage}[b]{1.0\linewidth}
  \centering
  \centerline{\includegraphics[width=14.5cm]{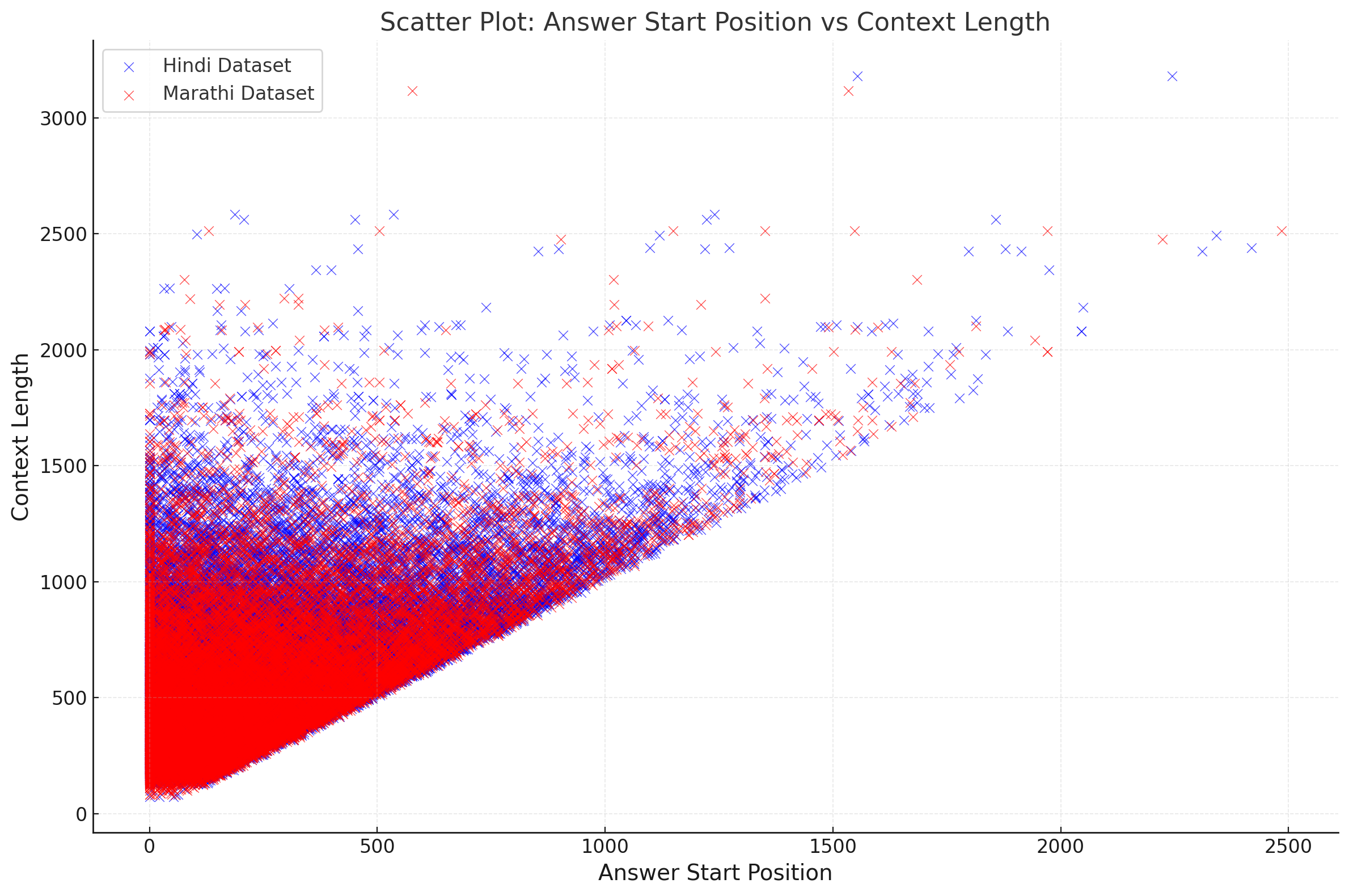}}
  \caption{\label{fig_dataStats} Answer Start Position vs. Context Length for Hindi (blue) and Marathi (red) datasets.}
\end{minipage}
\end{figure}

The datasets used in this study encompass diverse Indic languages, primarily focusing on Hindi and Marathi. The datasets are structured in a SQuAD-style format, comprising context paragraphs, questions, and corresponding answers. This standardized format ensures consistency across different QA tasks and facilitates effective model training and evaluation.\\
\\
The scatter plot in ~\cref{fig_scatterPlot} illustrates the relationship between question length and answer length across the Hindi (blue) and Marathi (red) datasets. The visualization highlights a clear positive correlation between the lengths of questions and their corresponding answers, with longer questions often leading to longer answers. Notably, the Marathi dataset shows slightly shorter average lengths compared to the Hindi dataset, reflecting language-specific characteristics. This analysis provides insights into the structural patterns within the dataset, aiding in understanding the data distribution and informing preprocessing decisions for the SSM-based QA model.
\\
\\
The scatter plot in ~\cref{fig_dataStats} illustrates the relationship between the starting position of answers within the context and the total context length for the Hindi (blue) and Marathi (red) datasets. The visualization shows that as the context length increases, the starting positions of answers become more varied, indicating that longer contexts are more likely to contain answers at different positions. This trend is consistent across both datasets, although the Marathi dataset tends to have shorter contexts on average compared to Hindi. This analysis highlights the diversity in answer positions relative to context length, which is crucial for designing models capable of handling varied input structures in question-answering tasks.
\newline
The correlation heatmaps provide insights into the relationships between key features in the Hindi and Marathi datasets, including context length, question length, answer length, and answer start position. As shown, context length exhibits a strong positive correlation with answer start position (0.48 for Hindi and 0.53 for Marathi), indicating that longer contexts tend to have answers starting further into the text. This relationship is more pronounced in the Marathi dataset. Additionally, the correlations between question length and answer length are negligible, suggesting that the lengths of questions and answers are largely independent. These observations reflect the structural differences in data organization and provide useful information for designing models that can effectively capture these patterns. The heatmaps highlight the similarities and differences in feature interactions between the two languages, guiding preprocessing and feature selection for downstream tasks.

\begin{figure}[H]
\begin{minipage}[b]{1.0\linewidth}
  \centering
  \centerline{\includegraphics[width=7.0cm]{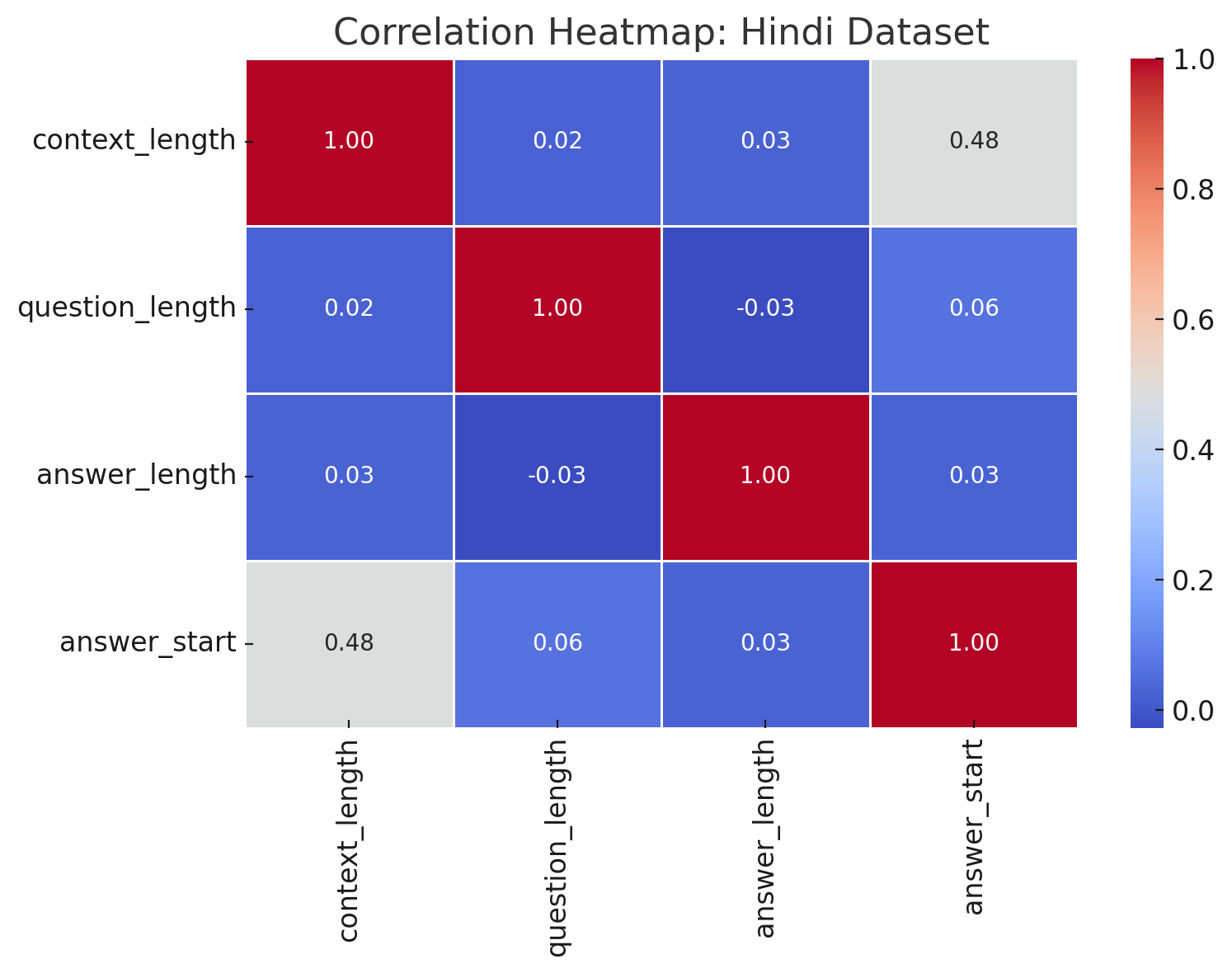}}
  \centerline{\includegraphics[width=7.0cm]{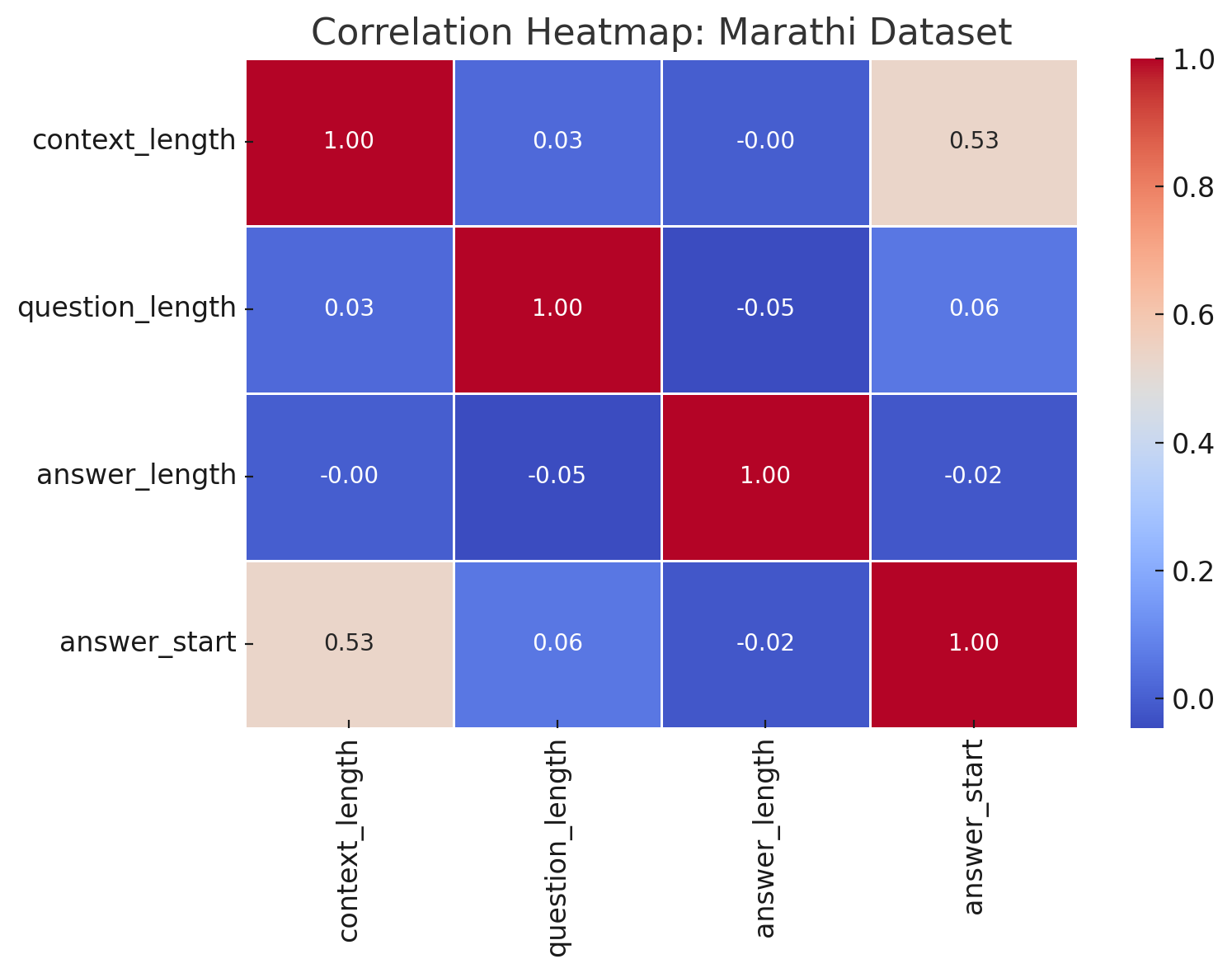}}
  \caption{\label{fig_pipeline} Correlation heatmaps for context length, question length, answer length, and answer start position in Hindi and Marathi datasets.}
\end{minipage}
\end{figure}
\section{Results}
\begin{figure}[htb]
\centering
\begin{minipage}[b]{1.0\linewidth}
  \centering
  \centerline{\includegraphics[width=8.0cm]{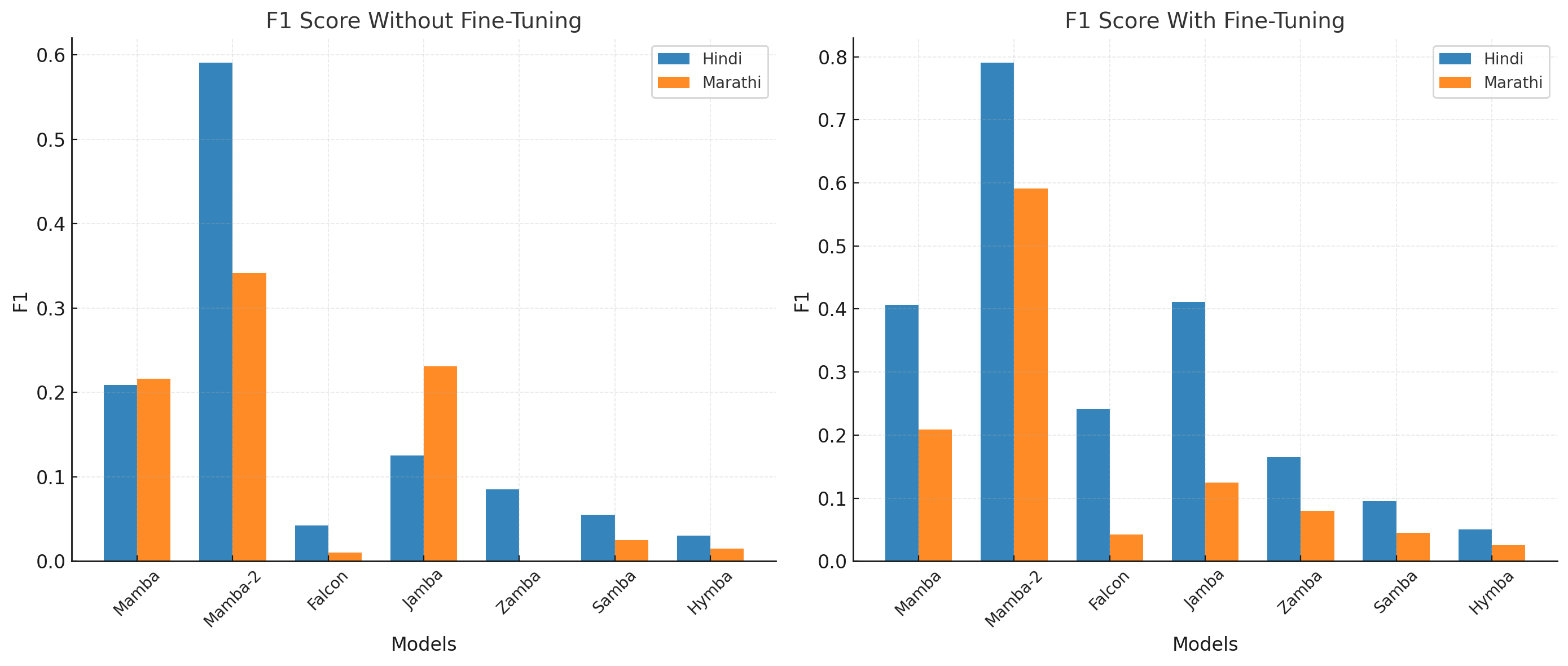}}
  \vspace{0.5cm} 
  \centerline{\includegraphics[width=8.0cm]{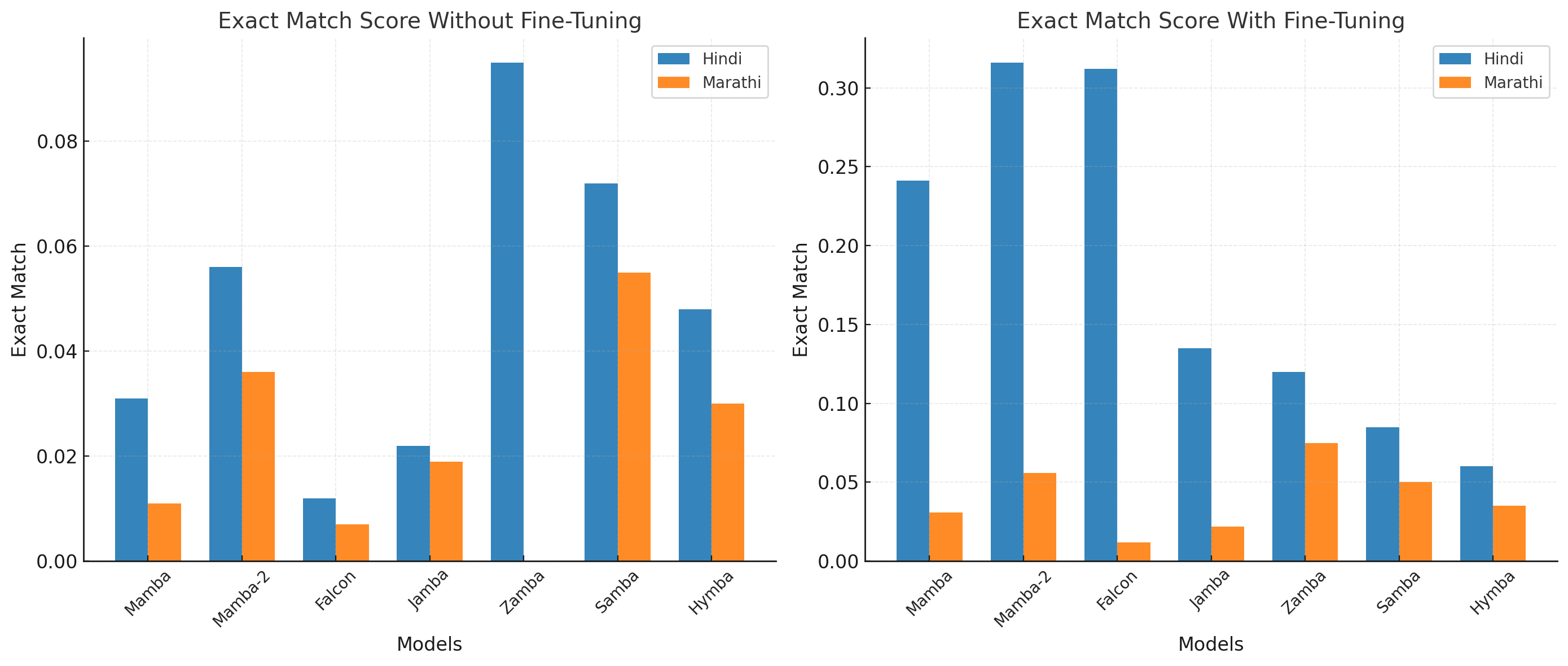}}
  \vspace{0.5cm}
  \centerline{\includegraphics[width=8.0cm]{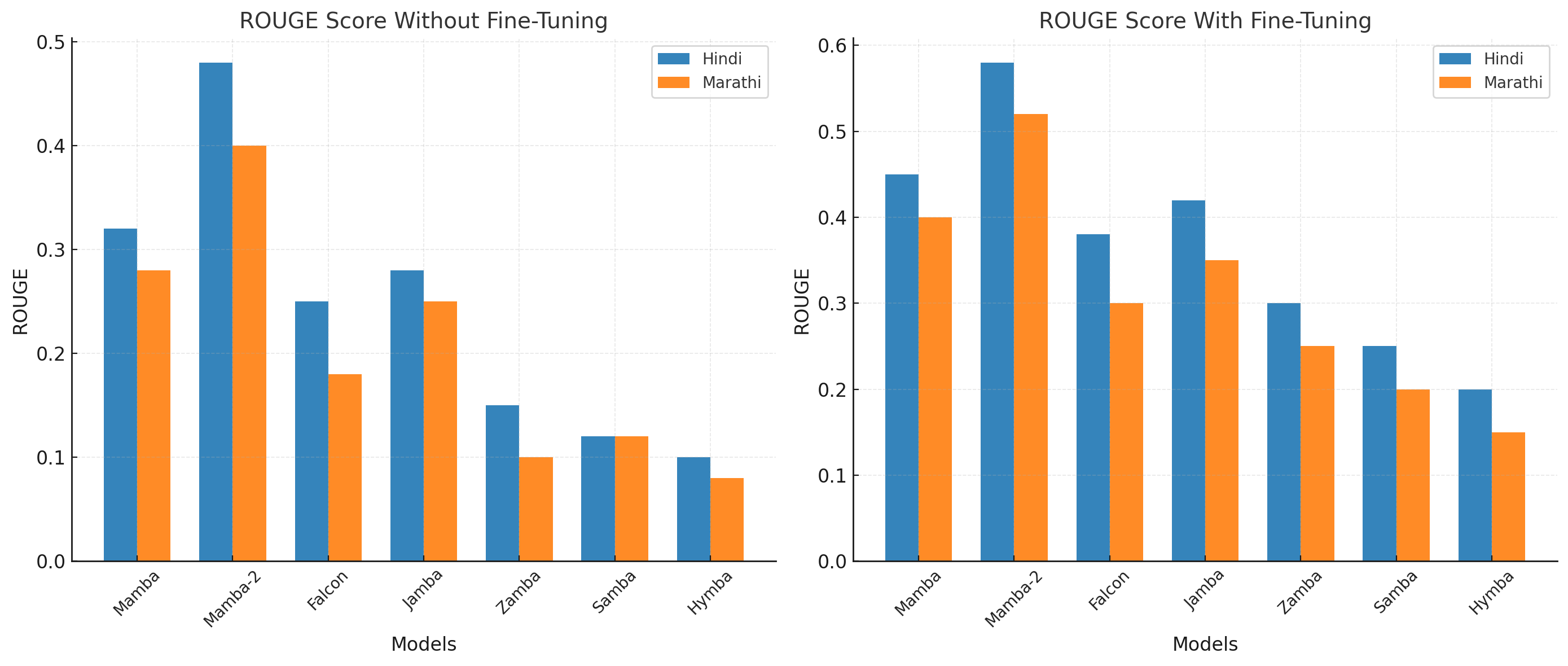}}
  \vspace{0.5cm}
  \centerline{\includegraphics[width=8.0cm]{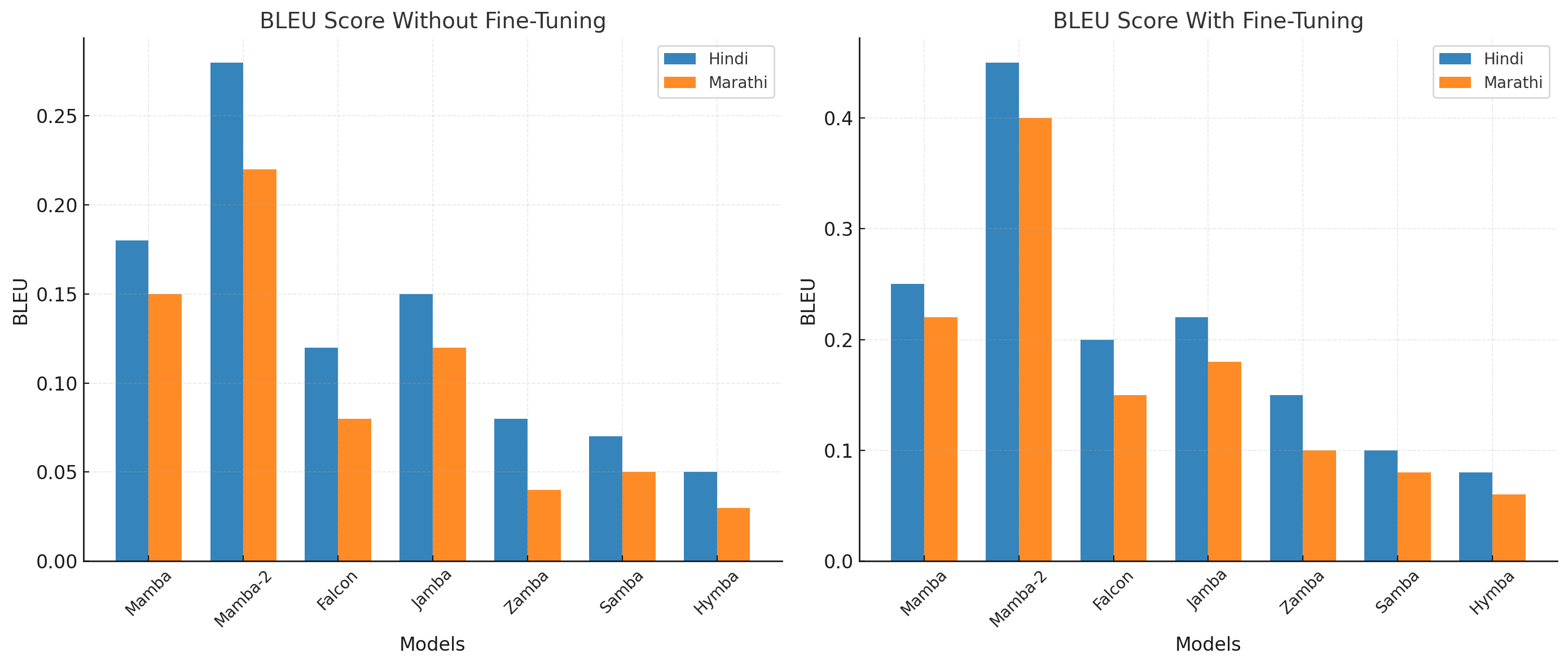}}
  \vspace{0.5cm}
  \centerline{\includegraphics[width=8.0cm]{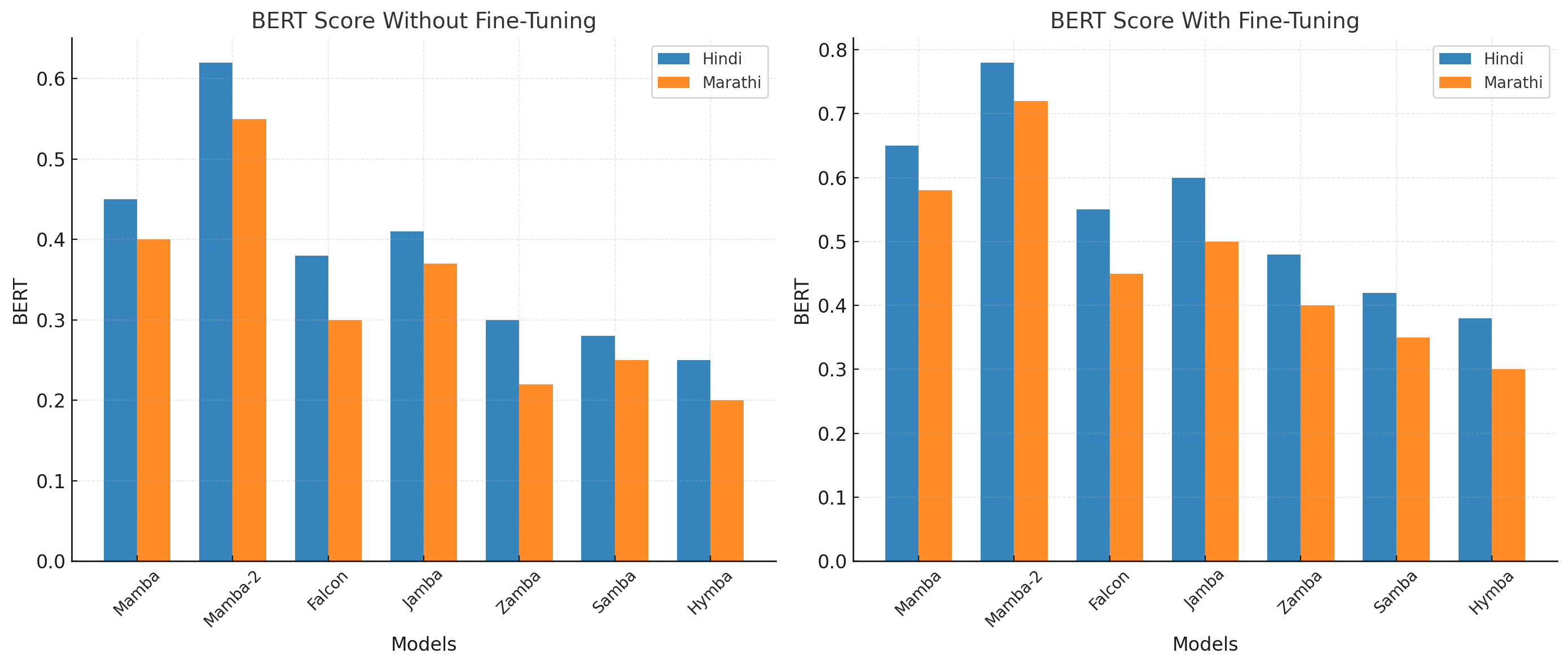}}
  \vspace{1.0cm} 
  \caption{\label{fig_results}Comparison of BLEU, ROUGE, F1, Exact Match, and BERT scores for Hindi and Marathi datasets before and after fine-tuning.}

\end{minipage}
\end{figure}

The results, as shown in ~\cref{fig_results}, illustrate the performance metrics for the Hindi and Marathi datasets, comparing outcomes before and after fine-tuning the model. Metrics such as BLEU, ROUGE, F1, Exact Match, and BERT scores are presented for both languages, highlighting the impact of fine-tuning on model performance. The left panel represent the scores without fine-tuning, while the right panels show the improved results after fine-tuning. Across both datasets, fine-tuning consistently improves all evaluation metrics, with notable gains in Exact Match and F1 scores, indicating better alignment between predicted and ground-truth answers. The BERT scores also demonstrate significant improvements, reflecting enhanced semantic understanding. These results emphasize the importance of fine-tuning for adapting the model to the nuances of Hindi and Marathi languages in the QA task.

\begin{figure*}[htb]
\begin{minipage}[b]{1.0\linewidth}
  \centering
  \centerline{\includegraphics[width=17.0cm]{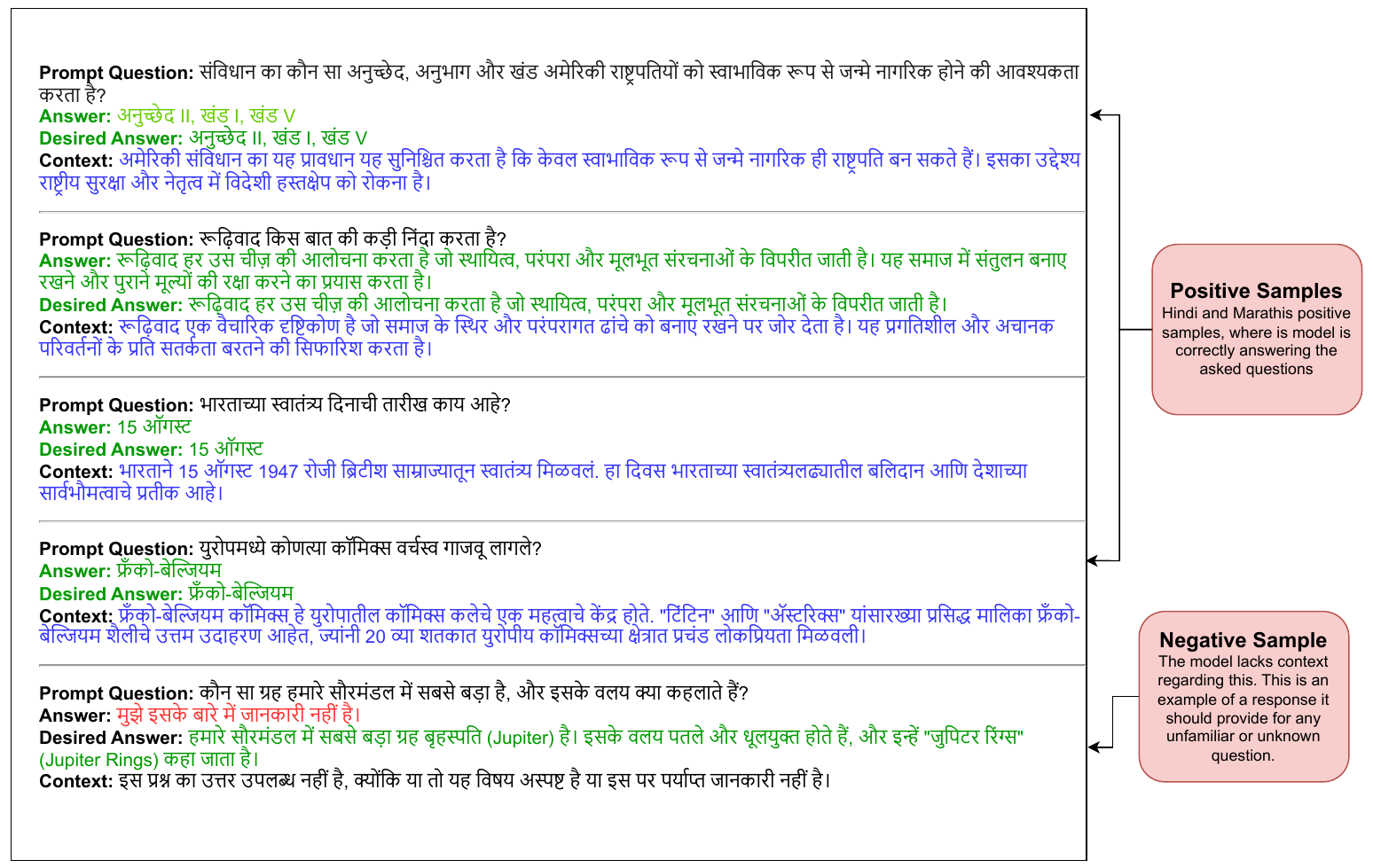}}
  \caption{\label{fig_sampleResults} Comparison of positive and negative samples for question-answering tasks, highlighting model responses.}
\end{minipage}
\end{figure*}

\end{document}